\documentclass[twocolumn]{article}

\usepackage{CJK}
\usepackage{enumerate}
\usepackage{amssymb}
\usepackage{amsmath}
\usepackage{dsfont}
\usepackage{indentfirst}
\usepackage{amsthm}
\usepackage{mathrsfs}
\usepackage{amsmath,amsfonts,graphicx}
\usepackage{amsmath}
\usepackage{amssymb}
\usepackage{graphicx}

\usepackage[numbers,sort&compress]{natbib}

\title{Discriminative Embedding Autoencoder with a Regressor Feedback for Zero-Shot Learning}
\author{Ying Shi}
\date{}
\begin{document}
\maketitle
\begin{abstract}
Zero-shot learning (ZSL) aims to recognize the novel object categories using the semantic representation of categories, and the key idea is to explore the knowledge of how the novel class is semantically related to the familiar classes. Some typical models are to learn the proper embedding between the image feature space and the semantic space, whilst it is important to learn discriminative features and comprise the coarse-to-fine image feature and semantic information. In this paper, we propose a discriminative embedding autoencoder with a regressor feedback model for ZSL. The encoder learns a mapping from the image feature space to the discriminative embedding space, which regulates both inter-class and intra-class distances between the learned features by a margin, making the learned features be discriminative for object recognition. The regressor feedback learns to map the reconstructed samples back to the the discriminative embedding and the semantic embedding, assisting the decoder to improve the quality of the samples and provide a generalization to the unseen classes. The proposed model is validated extensively on four benchmark datasets: SUN, CUB, AWA1, AWA2, the experiment results show that our proposed model outperforms the state-of-the-art models, and especially in the generalized zero-shot learning (GZSL), significant improvements are achieved.
\end{abstract}
\section{Introduction}
\label{sec:introduction}
Humans can distinguish approximately 30,000 basic object categories\cite{b1} and many more subordinate ones, e.g., breeds of dogs and many combination of attributes and objects. Importantly, humans are very good at recognizing objects without seeing any visual samples. In machine learning, this is considered as the problem of zero-shot learning (ZSL). ZSL has gained its popularity in object recognition task and can be used in a variety of research areas, such as neural decoding from fMRI images, face verification, object recognition, video understanding and natural language processing\cite{b2}. The traditional object recognition models are to predict the labels of object classes that already exist in the training set, however, zero-shot learning aims to build a model used to recognize object classes from a new category never seen before. Therefore, in the ZSL task, the seen classes in the training set and the unseen classes in the test set are disjoint. The main challenge of the zero-shot learning is how to generalize the models to identify the novel object classes without any labelled samples of these categories. Ideally, it would replicate the human ability to recognize objects from a few image or even from a semantic description\cite{b3}.

\begin{figure}
  \centering
  \includegraphics[scale=0.9]{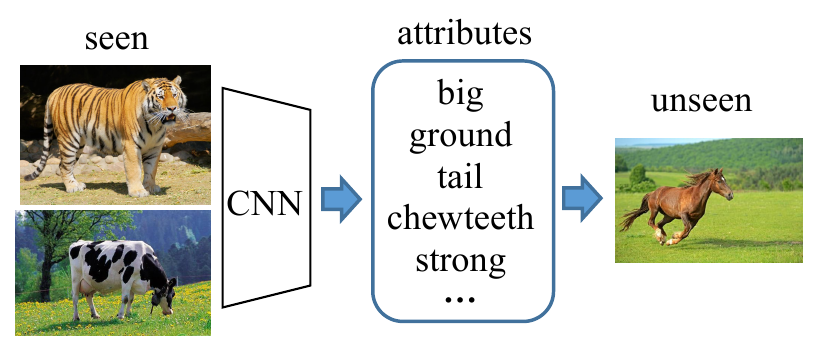}\\
  \caption{The illustration of zero-shot learning. Suppose the \emph{tiger} and the \emph{cow} are the training/seen classes, and the aim of zero-shot learning is to correctly classify a new/unseen class image such as the \emph{horse} by relating it to the images of seen class. The relationship between the classes can be provided by the semantic representation of categories such as \emph{attributes}.\label{img1}}
\end{figure}

The key idea of zero-shot learning is to explore the knowledge of how an unseen class is semantically related to the seen classes\cite{b4}. An example about ZSL is illustrated in Figure \ref{img1}. Seen and unseen classes are usually related in a high dimension vector space, called semantic space\cite{b5}, where the knowledge from seen classes can be transferred to unseen classes. The semantic representation of categories (e.g., semantic attribute annotations\cite{b6}, the text descriptions of the categories\cite{b7}, the semantic word vectors of the class names\cite{b8}, etc.) are required to share information between classes so that the knowledge learned from seen classes is transferred to unseen classes. Given a description of categories, each class name can be represented by an attribute vector or a semantic word vector. The semantic relationships between classes can be measured by a distance, e.g., the semantic of {\it zebra} and {\it horse} should be close to each other. One popular semantic representation is attributes, i.e., shared and nameable image properties of objects\cite{b4}. They are encoded in a high dimensional vector space. In this work, we focus on learning for ZSL with attributes.

Typically, some of the ZSL models learned a mapping function from an image feature space to a semantic embedding space using the labelled training data consisting of seen classes only; and then nearest neighbour (NN) search is performed in the projected semantic space where the label of the test image feature is matched by the nearest unseen class\cite{b5}. Existing models of ZSL focus on introducing linear or nonlinear mechanism and utilizing various optimization objective to learn the image feature-semantic mapping.

However, the final goal of ZSL is to classify the unseen classes. Therefore, the image feature and the semantic embedding should be discriminative to recognize different objects. Moreover, existing models mostly suffer from the projection domain shift problem\cite{b9}, that is, if the projection for the image feature is learned only from the seen classes, the projection of unseen class image features is likely to be shifted due to the bias of the seen classes. This shift could be far away from the accurate unseen classes\cite{b10}.

To address these issues, we propose a discriminative embedding autoencoder with a regressor feedback model for ZSL in both the image feature and semantic embedding space. Our contributions are three-fold:
\begin{itemize}
\item The discriminative embeddings cluster the intra-classes and separate the inter-classes by a margin, which preserve the discriminative information of the image features. An encoder acts as the discriminator.
\item The regressor feedback acts as the generator's regularizer to ensure the generated samples representative and accurate. The regressor feedback can assist the decoder to recover sufficient information contained in the image features and semantic embeddings to reconstruct the best image features.
\item Experimental results on public benchmark datasets validate the effectiveness of the proposed model, especially, the accuracy is significantly improved in the generalized zero-shot learning(GZSL).
\end{itemize}

The remainder of this paper is organized as follows. Section II summarizes the related work in zero-shot models, the autoencoder structure, and the generalized zero-shot learning. Section III introduces our proposed model architecture, our motivation, and every part of our model. Section IV describes the experiments, a comparison with existing methods, and analysis the performance of our model on the ZSL and GZSL settings. Finally, we conclude the paper with future work in Section V.

\section{Related Work}
Early work of zero-shot learning makes use of attribute with a two-stage approaches that first train different attribute classifiers and then recognize an image by comparing its predicted attributes with those of unseen classes\cite{b11}. For instance, DAP model\cite{b12} predicts the posterior of each attribute and then the class posteriors are calculated by maximizing a posterior. IAP model\cite{b12} first predicts the class posterior of seen classes, then the probability of each class is used to calculate the attribute posteriors of an image. In these methods, each attribute classifier is trained individually and the relationship between attributes for a class is not considered\cite{b11}.

\subsection{linear and nonlinear embedding models}
Recent advances in zero-shot learning typically learn an embedding from the image feature space to the semantic space, where the embedding is learned via a linear parameterized mapping. During testing, for an unseen class, the semantic vector is predicted and the neighbor class is assigned. The ALE\cite{b13} learns a bilinear compatibility function between the image and the attribute space using the ranking loss. The ESZSL\cite{b14} uses the square loss to learn the embedding and explicitly regularizes the objective. The SCoRe\cite{b3} adds a semantically consistent regularization to make the learned mapping perform better on test images. The SAE\cite{b10} uses a linear semantic autoencoder that its decoder acts as an additional constraint on the mapping to reconstruct the original image features.

In addition, non-linear compatibility mapping models have also been proposed. The LATEM\cite{b6} proposes piecewise compatibility modal learning which learns nonlinear compatibility function and the CMT\cite{b15} trains a neural network with two hidden layers to learn a nonlinear mapping from image feature space to word2vec space. The DEM\cite{b5} argues that the image feature space is more discriminative than semantic space, thus it proposes an end-to-end deep embedding model which maps from semantic space into the image feature space.

\subsection{embedding into common intermediate space}
Another direction of zero-shot learning embeds the image feature and the semantic into common intermediate space. The JLSE\cite{b16} maps the image features and the semantic space into two separate latent spaces, and measures their similarity by learning another bilinear compatibility function. The LAD\cite{b17} proposes to learn a latent attribute space, which is not only discriminative but also semantic-preserving. The SYNC\cite{b18} constructs the classifier of unseen classes by taking the linear combinations of base classifiers, which are trained in a discriminative learning framework. Annadani {\it et al.}\cite{b19} captures semantic relations defined on the categories themselves to learn the intermediate embedding space.

Different from them, our proposed model directly regulates both inter-class and intra-class distances between the learned features to achieve the discriminative embedding. The discriminative feature space and the semantic space jointly embed into the common intermediate space.

\subsection{generative models}
There are a few generative models that represent each class as a probability distribution. The GFZSL\cite{b20} treats each class-conditional distribution as a Gaussian and learns a regression function that maps a class embedding into the latent space. The GLAP\cite{b21} assumes that each class-conditional distribution follows a Gaussian and generates virtual samples of unseen classes from the learned distribution. Mukherjee {\it et al.}\cite{b22} learns a multimodal mapping where semantic and image embeddings of classes are both represented by Gaussian distributions. M. Bucher {\it et al.}\cite{b23} adopts generative model for data augmentation of unseen classes and uses these samples to train a classification model.

\subsection{the autoencoder structure}
The autoencoders are used for classification based on the assumption that higher dimensional features are better classification\cite{b24}. The SAE\cite{b10} model is a semantic autoencoder. Its decoder imposes an additional constraint in learning the visual to semantic mapping. This is very effective in mitigating the domain shift problem. This is because although the visual appearance of attributes may change from seen classes to unseen classes, the demand for more truthful reconstruction of the visual features is generalizable across seen and unseen domains, resulting in the learned project function less susceptible to domain shift\cite{b10}. Similarly, in our model, the encoder maps the image feature to the semantic embedding and the decoder reconstructs the original image feature to recover all the image feature and semantic information. Differently, our model proposes the discriminative feature in the embedding space and the decoder imposes a regressor feedback to the truthful and representative image feature. At test time, we use the decoder to generate the reconstructed unseen image features and then train an SVM classifier.

Zero-shot learning has been restrictive with a strong assumption that the image used to predict can only come from unseen classes. Therefore, generalized zero-shot learning has been proposed in \cite{b25} to generalize the zero-shot learning to the case where both seen and unseen classes are used during testing. Chao {\it et al.} \cite{b25} showed that it is nontrivial and ineffective to directly extend the current zero-shot learning approaches to solve the generalized zero-shot learning. Such a generalized setting, due to the more practical nature, is recommended as the evaluation settings for zero-shot learning \cite{b2}. We evaluate our model on the four benchmark datasets with SS and PS\cite{b4} for the two settings.

\section{Proposed Approach}
\subsection{Problem Definition}
In the zero-shot learning (ZSL), the set of train classes (also called seen classes) is defined as ${\cal S} \equiv \left\{ {\left( {x_i^s,y_i^s} \right)} \right\}_{i = 1}^{{n_s}}$, where $x_i^s \in {{\cal X}_{\cal S}}$ is the {\it i}-th image feature of the seen class and $y_i^s \in {{\cal Y}_{\cal S}}$ is its corresponding class label, ${{n_s}}$ represents the number of the image feature of seen classes. The set of test classes (also called unseen classes) is defined as ${\cal U} \equiv \left\{ {\left( {x_j^u,y_j^u} \right)} \right\}_{j = 1}^{{n_u}}$, where $x_j^u \in {{\cal X}_{\cal U}}$ is the {\it j}-th image feature of the unseen class and $y_j^u \in {{\cal Y}_{\cal U}}$ is the label of it, ${{n_u}}$ represents the number of the image feature of unseen classes. The seen and unseen classes are disjoint, i.e., ${{\cal Y}_{\cal S}} \cap {{\cal Y}_{\cal U}} = \emptyset$. We work in the image feature space instead of the image space. A key of zero-shot learning is the semantic embedding of the class labels. In this work, the class semantic embeddings are represented to the attribute vectors. Distributed word representations of the class name such as {\it word}2{\it vec} \cite{b26} have also been used as the semantic embedding. The attributes for both seen and unseen classes can be denoted as ${{\cal A}_{\cal S}} \equiv \left\{ {a_i^s} \right\}_{i = 1}^{{c_s}}$ and ${{\cal A}_{\cal U}} \equiv \left\{ {a_j^u} \right\}_{j = 1}^{{c_u}}$, where $a_i^s$ and $a_j^u$ respectively indicate the attribute vectors for the {\it i}-th seen class and the {\it j}-th unseen class, ${{c_s}}$ and ${{c_u}}$ represent the number of the attribute vectors for seen classes and unseen classes, respectively. At test time, given a test image feature ${x^u}$ and the attribute of test classes $a^u$, the goal of ZSL is to predict the correct class of ${x^u}$, without trained classifier by unseen classes.

\subsection{Model Architecture}

\begin{figure*}
  \centering
  \includegraphics[scale=0.5]{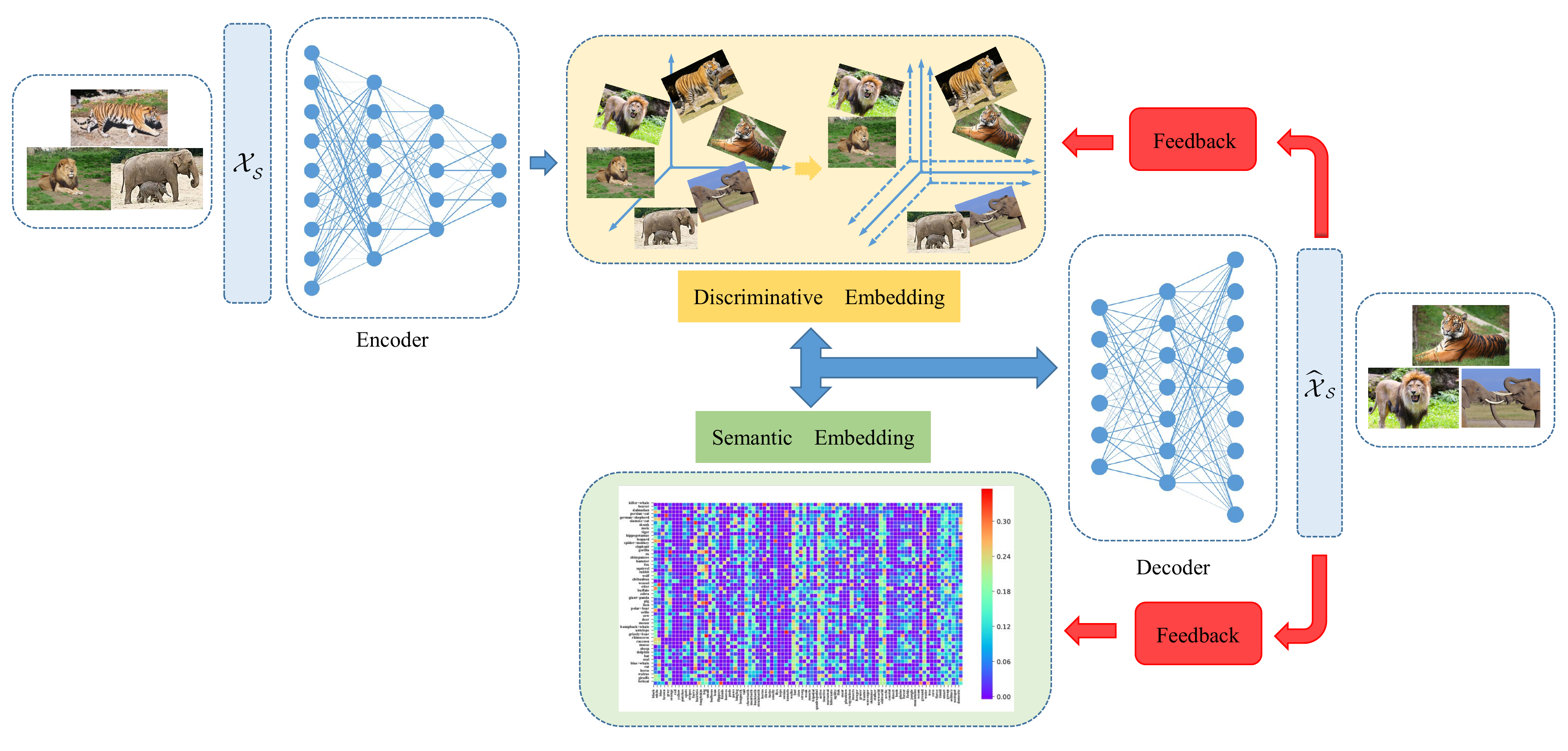}\\
  \caption{The framework of our proposed model.\label{fig1}}
\end{figure*}

The framework of our model is shown in Figure \ref{fig1}. Our model consists of four different components: 1) The image features ${{\cal X}_{\cal S}}$ are encoded to the discriminative embeddings which have the same dimension as the semantic embeddings ${{\cal A}_{\cal S}}$. 2) The discriminative embeddings and the semantic embeddings ${{\cal A}_{\cal S}}$ are concatenated and decoded to reconstruct the original image features ${{\cal X}_{\cal S}}$. 3) The reconstructed image features ${\widehat{\cal X}_{\cal S}}$ are mapped back to the corresponding semantic embeddings ${{\cal A}_{\cal S}}$ and the discriminative embeddings, providing a feedback to the decoder. 4) Inputting each unseen class vector $a_j^u$ to the decoder can generate the reconstructed unseen classes data used for classification of the unseen classes. An autoencoder is one realisation of the encoder-decoder paradigm. In our model, the encoder acts as a discriminator, and the decoder acts as a generator. The autoencoder is responsible for generating the image features. The more truthful reconstructed image feature is generalizable across the seen classes ${{\cal Y}_{\cal S}}$ and the unseen classes ${{\cal Y}_{\cal U}}$, which builds a bridge between classes.

\subsection{Motivation}
The autoencoder aims to get the reconstructed image features that are expected to recover sufficient semantic information and the discriminative features, and then provides a good generalization to the unseen classes. For this goal, we propose the discriminative embedding and the regressor feedback, and details of them are in the following two sections. On one hand, the discriminative embeddings have learned the discriminative features by a nonlinear dense network with the triplet loss\cite{b27}, and the learned features preserve the discriminative information. On the other hand, the discriminative embeddings and semantic embeddings are concatenated to train the generator. The regressor feedback acts as the generator's regularizer from the output of the generator back to the semantic embeddings and the discriminative embeddings, which makes the generation of samples contain sufficient discriminative and semantic information. The recurrent structure achieves the coarse-to-fine process at each iteration. The generator is used for generating the unseen image feature to train a classifier.

In other words, the discriminative embeddings and semantic embeddings may have a correlation relationship. Fusing them can help the generator reconstruct the image features that are representative and accurate for the corresponding class, and then can transfer the relationship knowledge from the seen classes to unseen classes. Also, that's why two same unseen semantic embeddings are concatenated during testing, which is valid to act as the input of the generator.

\subsection{Encoder}
The image features ${{\cal X}_{\cal S}}$ is trained by the nonlinear dense network to obtain the discriminative embeddings:
\begin{equation}
{\phi _e}\left( {x_i^s} \right) = W_e^T*x_i^s,\quad x_i^s \in {{\cal X}_{\cal S}},
\end{equation}
 where ${\phi _e}\left( {x_i^s} \right)$ is the discriminative embedding and the output of the last dense layers, $*$ denotes a set of operations of the encoder and ${{W_e}}$ indicates the overall parameters of the encoder. The image feature $x_i^s$ is projected into the discriminative embedding ${\phi _e}\left( {x_i^s} \right)$, which generates discriminative image features.

The image features ${{\cal X}_{\cal S}}$ as the input of the encoder pass through two hidden layers, followed by a dense layer with the linear activation. The discriminative embeddings ${\phi _e}\left( {{{\cal X}_{\cal S}}} \right)$ as the output have the same dimension as the semantic embedding vector.

\subsection{Discriminative Embedding}
Most of the embedding models to solve the ZSL problem are based on the semantic embeddings. However, it is of limited size and not discriminative. To address this issue, we introduce the discriminative embeddings, which learn the image features from high dimension to low dimension and make the learned features be discriminative for object recognition.

We want to ensure that an image feature $x_i^s$ (anchor) of a specific class is closer to all other image feature $x_k^s$ (positive) of the same class than any image feature $x_j^s$ (negative) of any other class. The triplet loss\cite{b27} minimizes the distance between an anchor and a positive, both of which have the same identity, and maximizes the distance between the anchor and a negative of a different identity\cite{b28}. We utilize the triplet loss to learn the discriminative embeddings with regulating the inter and intra class distances between the learned features:
\begin{equation}
{\cal L} = \frac{1}{{{n_s}}}\sum\limits_{i = 1}^{{n_s}} {\max \left( {0,m + {d_1} - {d_2}} \right)},
\end{equation}
where ${d_1} = d\left( {{\phi _e}\left( {x_i^s} \right),{\phi _e}\left( {x_k^s} \right)} \right)$, ${{\phi _e}\left( {x_i^s} \right)}$ and ${{\phi _e}\left( {x_k^s} \right)}$ are the learned image features from the same class. ${d_2} = d\left( {{\phi _e}\left( {x_i^s} \right),{\phi _e}\left( {x_j^s} \right)} \right)$, ${{\phi _e}\left( {x_i^s} \right)}$ and ${{\phi _e}\left( {x_j^s} \right)}$ are from different classes. $d\left( {x,y} \right)$ is the squared Euclidean distance between $x$ and $y$. The distance between the inter/intra class should be lager than a margin $m > 0$.

In the embedding space, the semantic embeddings ${{\cal A}_{\cal S}}$ and the discriminative embeddings are concatenated, which allows the generator learned from not only the semantic embeddings ${{\cal A}_{\cal S}}$ but also the image features ${{\cal X}_{\cal S}}$. They contain meaningful and complementary information, and fusing them can potentially improve the quality of the reconstructed image features.

\subsection{Decoder}
The decoder acts as the generator that the mapping must be able to reconstruct the original image features. It is expected to preserve sufficient semantic and discriminative information so as to reconstruct the high-quality and class-specific image features. Specifically, the discriminative embeddings ${\phi _e}\left( {{{\cal X}_{\cal S}}} \right)$ concatenated with the semantic embeddings ${{\cal A}_{\cal S}}$ are projected to the image features ${{\cal X}_{\cal S}}$. The reconstructed image feature denotes:
\begin{equation}
\widehat{x}_i^s = {\phi _d}\left( {\left[ {\phi _e}\left( {x_i^s} \right),a_i^s\right];{W_d}} \right),\quad\widehat{x}_i^s \in {\widehat{\cal X}_{\cal S}},
\end{equation}
 where $\widehat{x}_i^s$ denotes the reconstructed image feature of the {\it i}-th seen class, ${\phi _d}$ denotes the output of the last dense layers and ${{W_d}}$ indicates the overall parameters of the decoder. ${\widehat{\cal X}_{\cal S}}$ denotes the reconstructed image feature space.

We observed that training with two hidden layers in the decoder quickly overfits to the seen classes, then one hidden layer is used with the LeakyReLU \cite{b30} activation, followed by a dense layer with the linear activation. The input is the semantic embedding space ${{\cal A}_{\cal S}}$ and the discriminative embeddings ${\phi _e}\left( {{{\cal X}_{\cal S}}} \right)$, and the output is the reconstructed image feature space ${\widehat{\cal X}_{\cal S}}$.

The encoder and the decoder are linked together by the discriminative embeddings. During training, the image feature ${x_i^s}$ is the input of the encoder and the reconstructed image feature ${\widehat{x}_i^s}$ is the decoder's output, and the semantic embedding $a_i^s$ is the intermediate condition. the reconstruction objective function becomes:
\begin{equation}
{{\cal L}_{reconstr}}\left( { x_i^s,a_i^s;\phi ,W} \right) = \frac{1}{{{n_s}}}{\sum\limits_{i = 1}^{{n_s}} {\left\| {x_i^s - \widehat{x}_i^s} \right\|} ^2},
\end{equation}
where $\phi$ denotes the mapping from ${x_i^s}$ to ${\widehat{x}_i^s}$ and $W$ is the overall parameters of the decoder. It is necessary that the output of the decoder can reconstruct the image feature.

\subsection{Regressor Feedback}

\begin{figure}
  \centering
  \includegraphics[scale=0.8]{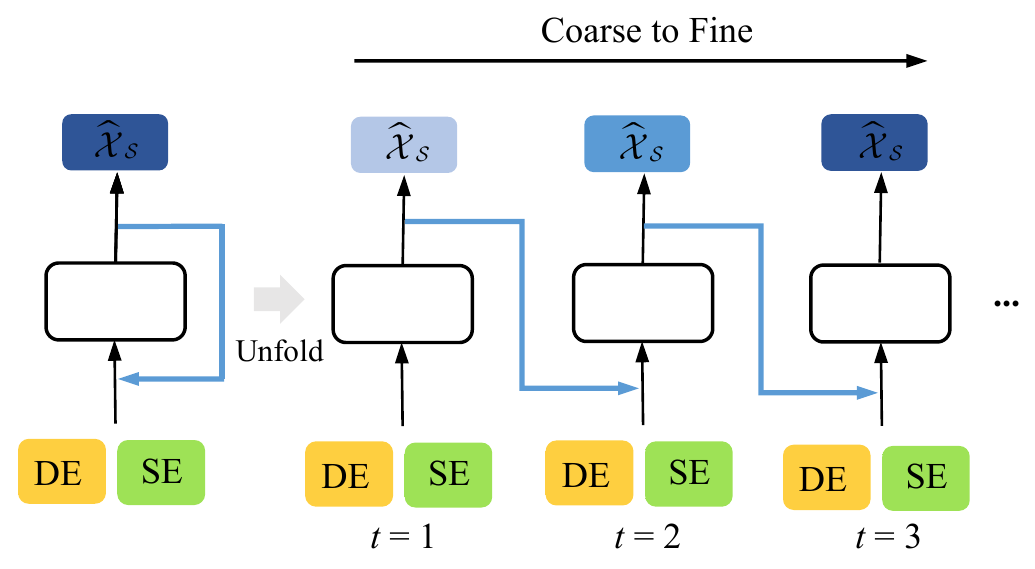}\\
  \caption{The feedback mechanism in learning generator. Blue arrows represent the feedback connections. DE is discriminative embeddings, SE is semantic embeddings, and \emph{t} is the iteration times.\label{fig3}}
\end{figure}

The feedback mechanism\cite{b29} allows the network to carry high-level information back to previous layers and refine low-level encoded information. It reroutes the output back into the model to produce the best output in each iteration. Motivated by this, we apply the feedback mechanism to our model architectures, as shown in Figure \ref{fig3}.

Our model consists of a mapping from the decoder's output ${\widehat{\cal X}_{\cal S}}$ to the semantic embeddings ${{\cal A}_{\cal S}}$ and the discriminative embeddings. The output of the decoder at each iteration flows into the next iteration to modulate the input. This mapping is a multivariate regression network learned jointly with the rest of the model and not an independent part. The recurrent structure is trained to produce better unseen classes at each iteration, i.e., coarse samples which involve fewer features of the class are generated in the first iterations and finer ones can be achieved as the proceeding of the iterations. The generator with the feedback results in generation of samples that can be discriminated easily.

\textbf{Regressor Feedback Network} The regressor maps the generator back to the semantic embeddings and the discriminative embeddings, ensuring that the generator can recover sufficient discriminative and semantic information and provide a generalization to the unseen classes. Following the output of the generator/decoder, the reconstructed image features ${\widehat{\cal X}_{\cal S}}$ enter one hidden layer through a feedback connection. We use the regressor to improve the generator, which is learned using two source of data. The reconstructed image feature ${\widehat{x}_i^s}$ maps to the sematic embedding, and we can define a sematic loss, given by:
\begin{equation}
{{\cal L}_{reg\_sem}} = \frac{1}{{{n_s}}}{\sum\limits_{i = 1}^{{n_s}} {\left\| {a_i^s - {\phi _{reg}}\left( {\widehat{x}_i^s} \right)} \right\|} ^2},
\end{equation}
where ${{\phi _{reg}}}$ represents the regressor mapping, i.e., the feedback connection.
The reconstructed image feature ${\widehat{x}_i^s}$ maps to the discriminative embedding, and we can define a discriminative loss, given by:
\begin{equation}
{{\cal L}_{reg\_dis}} = \frac{1}{{{n_s}}}{\sum\limits_{i = 1}^{{n_s}} {\left\| {{\phi _e}\left( {x_i^s} \right) - {\phi _{reg}}\left( {\widehat{x}_i^s} \right)} \right\|} ^2}.
\end{equation}
The overall training objective of the regressor is defined as the following weighted combination of the above two:
\begin{equation}
{{\cal L}_{reg}} = {{\cal L}_{reg\_sem}} + \lambda {{\cal L}_{reg\_dis}},
\end{equation}
where $\lambda$ is a weighting coefficient that controls the importance of first and second terms. The feedback mechanism allows the network to carry the ${\widehat{x}_i^s}$ back to previous layers and refine the ${a_i^s}$ and ${{\phi _e}\left( {x_i^s} \right)}$ encoded information. It reroutes the output back into the model to improve the quality of the reconstructed data that can be used to train the final classifier.

\subsection{Full Obiective}
Combining the objective functions introduced above, the full objective of our proposed model is:
\begin{equation}
{\cal L} = {{\cal L}_{encoder}} + \alpha {{\cal L}_{reconstr}} + \beta {{\cal L}_{reg}},
\end{equation}
where $\alpha$ and $\beta$ are trade-off parameters for different objectives. We minimise the objective function to estimate the parameters of our model.

The encoder preserves the discriminative information of the image feature and the decoder with an additional regressor regularizer generates data that are highly discriminative in nature, as guided by the semantic embeddings and the discriminative embeddings. The autoencoder mechanism aims to generate the reconstructed data containing sufficient semantic and discriminative information that can provide a generalization to unseen classes.

\subsection{ZSL prediction}
For the unseen classes ${a_j^u}$, their semantic embedding vectors are known. For classification, the generator of our trained model is used to map the semantic embedding of the unseen class to generate the corresponding reconstructed unseen image feature:
\begin{equation}
\widehat{x}_j^u = {\phi _d}\left( {a_j^u;{W_d}} \right),\quad\widehat{x}_j^u \in {\widehat{\cal X}_{\cal U}},
\end{equation}
where ${\widehat{\cal X}_{\cal U}}$ represents the reconstructed unseen image feature space. Once the data is generated, we can use the reconstructed unseen image feature $\widehat{x}_j^u$ and its label $y_j^u$ to train the classifier for the unseen classes. In this work, an SVM classifier and the accuracy score are used to predict an unseen class label:
\begin{equation}
\widehat{y} = {\phi _{cls}\left(\widehat{x}_j^u,x_j^u,y_j^u \right)},
\end{equation}
where ${{\phi _{cls}}}$ denotes the output of the SVM classifier. Finally, the most matched unseen class is selected.

\section{Experiments}
\subsection{Datasets}
Our proposed model is evaluated on ZSL benchmark datasets: SUN Attribute (SUN)\cite{b31}, Caltech-UCSD Birds 200-2011 (CUB)\cite{b32}, Animals with Attributes 1 (AWA1)\cite{b33}, Animals with Attributes 2 (AWA2)\cite{b2}. Details of these datasets are listed in Table \ref{tab1}.
\begin{table*}
\centering
\caption{Details of dataset statistics for SUN, CUB, AWA1 and AWA2 in terms of granularity, number of attributes, number of classes in ${{\cal Y}_{\cal S}}$ and ${{\cal Y}_{\cal U}}$, number of images for SS and PS.}
\begin{tabular}{cccc|ccccc}
  \hline
    &  &  &  &  & \multicolumn{2}{c}{\textbf{At Training Time}} & \multicolumn{2}{c}{\textbf{At Testing Time}} \\
  \textbf{Datasets}  & \textbf{Granularity} & \textbf{Att} & \textbf{${{\cal Y}_{\cal S}}/{{\cal Y}_{\cal U}}$} & \textbf{Total} &\textbf{SS(${{\cal Y}_{\cal S}}$)} & \textbf{PS(${{\cal Y}_{\cal S}}$)} & \textbf{SS(${{\cal Y}_{\cal U}}$)} & \textbf{PS(${{\cal Y}_{\cal S}}/{{\cal Y}_{\cal U}}$)} \\
  \hline
  SUN&fine&102&645/72& 14340 & 12900 & 10320 &1440 & 2580/1440 \\
  CUB&fine&312&150/50& 11788 & 8855 & 7057 & 2933 &	1764/2967 \\
  AWA1&coarse&85&40/10&	30475 &	24295 &	19832 &	6180 & 4958/5685 \\
  AWA2&coarse&85&40/10&	37322 &	30337 &	23527 &	5985 & 5882/7913 \\
  \hline
\end{tabular}
\label{tab1}
\end{table*}

\begin{table*}
\centering
\caption{Zero-Shot Learning results on SUN, CUB, AWA1 and AWA2 using SS and PS with ResNet features. The results report top-1 accuracy in \%.}
\begin{tabular}{ccc|cc|cc|cc}
  \hline
 &\multicolumn{2}{c|}{\textbf{SUN}}&\multicolumn{2}{c|}{\textbf{CUB}}&\multicolumn{2}{c|}{\textbf{AWA1}}&\multicolumn{2}{c}{\textbf{AWA2}}\\
  \textbf{Method}&\textbf{SS}&\textbf{PS}&\textbf{SS}&\textbf{PS}&\textbf{SS}&\textbf{PS}&\textbf{SS}&\textbf{PS}\\
  \hline
  DAP\cite{b12}&38.9&39.9&37.5&40.0&57.1&44.1&58.7&46.1\\
  IAP\cite{b12}&17.4&19.4&27.1&24.0&48.1&35.9&46.9&35.9\\
  CONSE\cite{b8}&44.2&38.8&36.7&34.3&63.6&45.6&67.9&44.5\\
  CMT\cite{b15}&41.9&39.9&37.3&34.6&58.9&39.5&66.3&37.9\\
  SSE\cite{b16}&54.5&51.5&43.7&43.9&68.8&60.1&67.5&61.0\\
  LATEM\cite{b6}&56.9&55.3&49.4&49.3&74.8&55.1&68.7&55.8\\
  ALE\cite{b13}&59.1&58.1&53.2&54.9&78.6&59.9&80.3&62.5\\
  DEVISE\cite{b35}&57.5&56.5&53.2&52.0&72.9&54.2&68.6&59.7\\
  SJE\cite{b36}&57.1&53.7&55.3&53.9&76.7&65.6&69.5&61.9\\
  ESZSL\cite{b14}&57.3&54.5&55.1&53.9&74.7&58.2&75.6&58.6\\
  SYNC\cite{b18}&59.1&56.3&54.1&\textbf{55.6}&72.2&54.0&71.2&46.6\\
  SAE\cite{b10}&42.4&40.3&55.8&33.3&80.7&53.0&80.8&54.1\\
  \hline
  \textbf{Ours}&\textbf{64.3}&\textbf{62.4}&\textbf{56.1}&53.9&\textbf{81.0}&\textbf{79.8}&\textbf{81.2}&\textbf{78.5}\\
  \hline
\end{tabular}
\label{tab2}
\end{table*}

\begin{figure*}
  \centering
  \includegraphics[scale=0.4]{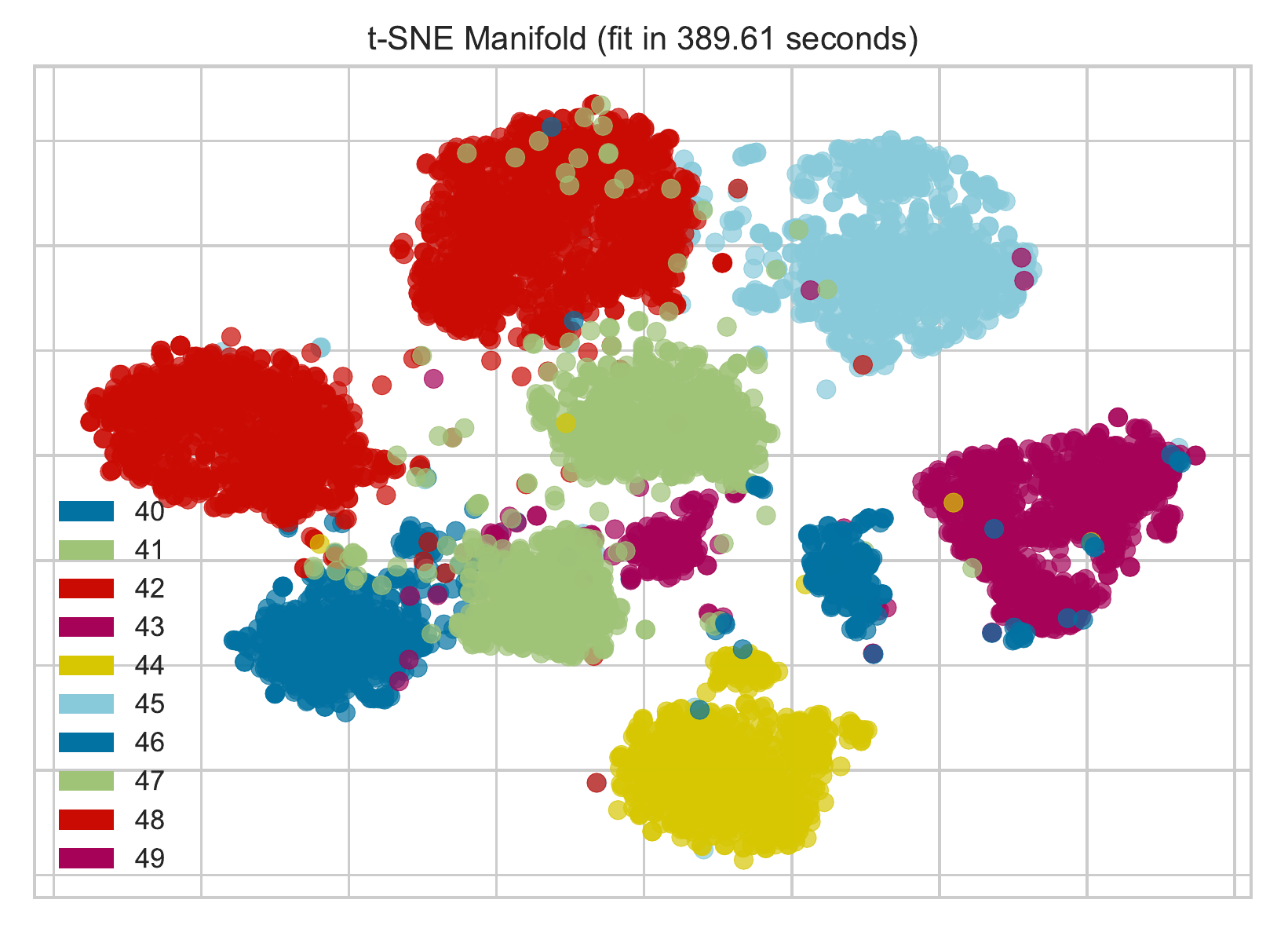}
  \includegraphics[scale=0.4]{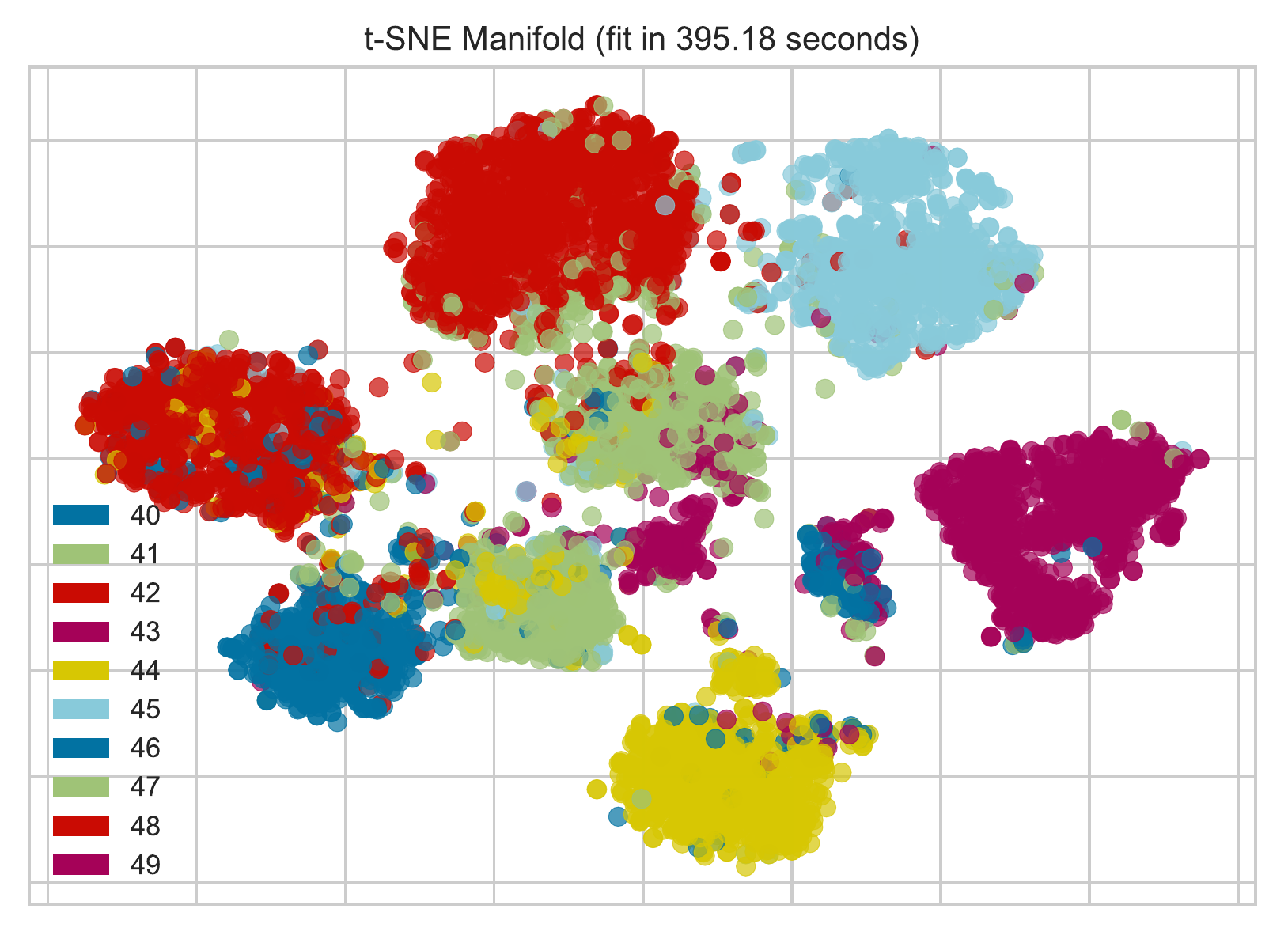}\\
  \caption{t-SNE visualizations of the 10 unseen test classes of AWA2 dataset. The left part shows the original test image features and the right part shows the predicted test image features. The numbers 40-49 correspond to the 10 unseen class labels: 40: sheep, 41: dolphin, 42: bat, 43: seal, 44: blue+whale, 45: rat, 46: horse, 47: walrus, 48: giraffe, 49: bobcat.}\label{tsne}
\end{figure*}

It is observed in \cite{b4} that some of the testing classes in the standard splits (SS) of the datasets are the subsets of the Imagenet\cite{b34} classes. Hence, extracting features from Imagenet trained model will not represent a true performance, but the SS has been widely used by some recent zero-shot learning models. Xian {\it et al}. \cite{b4} propose a new dataset split - the propose split (PS) ensuring that none of the test classes contain ImageNet classes. The differences between SS and PS are shown in Table \ref{tab1}.

The image features are 2048-dim top-layer pooling units of the 101-layered ResNet that is pre-trained on ImageNet\cite{b4}. All methods are evaluated with published image features. Continuous values between 0 and 1 are used for the class attributes that are provided with the datasets as the semantic embeddings which perform better than the binary attributes\cite{b4}. The semantic embedding is a {\it word}2{\it vec} trained on Wikipedia provided by \cite{b18}. We use the average per class top-1 accuracy\cite{b4} as the evaluation criteria. It is defined as follows:
\begin{equation}
acc_{avg}^{per - class} = \frac{1}{{\left\| {\cal Y} \right\|}}\sum\limits_{j = 0}^{\cal Y} {\frac{{N_{correct}^{\left( {class - i} \right)}}}{{N_{total}^{\left( {class - i} \right)}}}},
\end{equation}
where ${N_{correct}^{\left( {class - i} \right)}}$ and ${N_{total}^{\left( {class - i} \right)}}$ represent the correct number of predictions for the {\it i}-th class and the total number of the {\it i}-th class respectively.

\textbf{Implementation Details} Our proposed model is composed of 8 dense layers with output channel numbers as $2048 \to 1024 \to 512 \to D+D \to 1024 \to 2048 \to 1024 \to D+D$, where D represents dimension of the semantic embedding vector and the last two layers represent the feedback network. All activation functions are LeakyReLU \cite{b39} with the negative slope of 0.2, except the output of the encoder, the decoder and the regressor which are linear. The mean square error loss is used to reduce the discrepancy between the vectors.

For the discriminative embeddings, it is crucial to select hard triplets, that are active and can contribute to improve the model. We choose the strategy to train the triplet loss\cite{b28}: for each batch, the first step is to classify the positive/same classes and the negetive/different classes, and then each image feature is considered as the anchor and its hardest positive image feature such that $\max {d_1}$ and hardest negative image feature such that $\min {d_2}$ are selected to calculate the loss. The discriminative embeddings and the semantic embeddings have the same dimension, and they jointly embed into the intermediate embedding space, that is, the input dimension of the generator is 2 times of the semantic embedding dimension.

\subsection{Experimental Results}
The results of the zero-shot learning experiment are given in Table 2. The SS and PS are used in SUN, CUB, AWA1 and AWA2 datasets to achieve the accuracy of each class (top-1 accuracy). Compared with other 12 methods, our proposed model achieves an improvement over state of the art. The SS and PS of the AWA1 are increased by 0.3\% and 14.2\%, and the SS and PS of the AWA2 are increased by 0.4\% and 16.0\%, respectively. The SS and PS of SUN are increased by 5.2\% and 4.3\%. The SS of CUB is increased by 0.3\%. We may note that currently no other single method claims the best results on all the datasets simultaneously.

The experiment results show that our proposed model performs better in the coarse-grained datasets (AWA-1, AWA-2) than in the fine-grained datasets (SUN, CUB). The discriminative embedding has more effect on the coarse-grained datasets where the inter-classes semantics are much different, because the triplet loss minimizes the pairwise distances between all similarly labeled examples and separates examples from different classes by a large margin\cite{b27}, making the learned features be discriminative for object recognition. At the same time, the regressor feedback helps the generator samples be representative and accurate on the corresponding class and achieves the coarse-to-fine process at each iteration, especially it contributes to the fine-grained datasets where the intra-classes have complex semantics. Besides, the large number of classes and relatively fewer training samples in the SUN and CUB make the accuracy improve slightly.

We perform t-SNE visualization\cite{b37} to compare the test image features predicted by our proposed model (left) and the original test image features for the AWA2 dataset (right) in Figure \ref{tsne}. Each color represents clustering in the same class and all the image features embed into two dimensions using t-SNE. Compared with the true data, the predicted image features are close to the original ones for most classes, which indicates that our proposed model is able to capture the underlying distribution and performs better on the dataset.
\begin{figure*}
  \centering
  \includegraphics[scale=0.4]{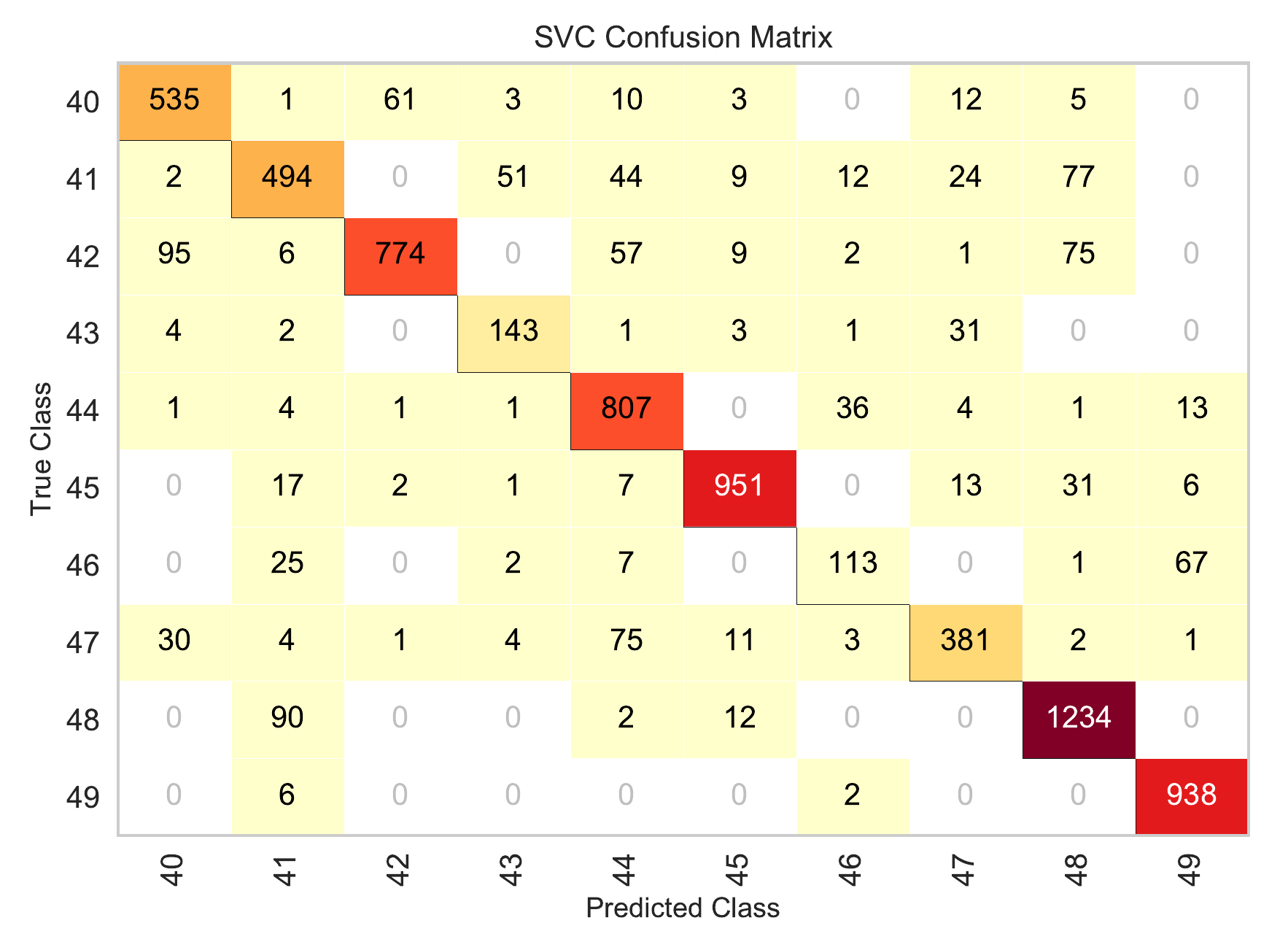}
  \includegraphics[scale=0.4]{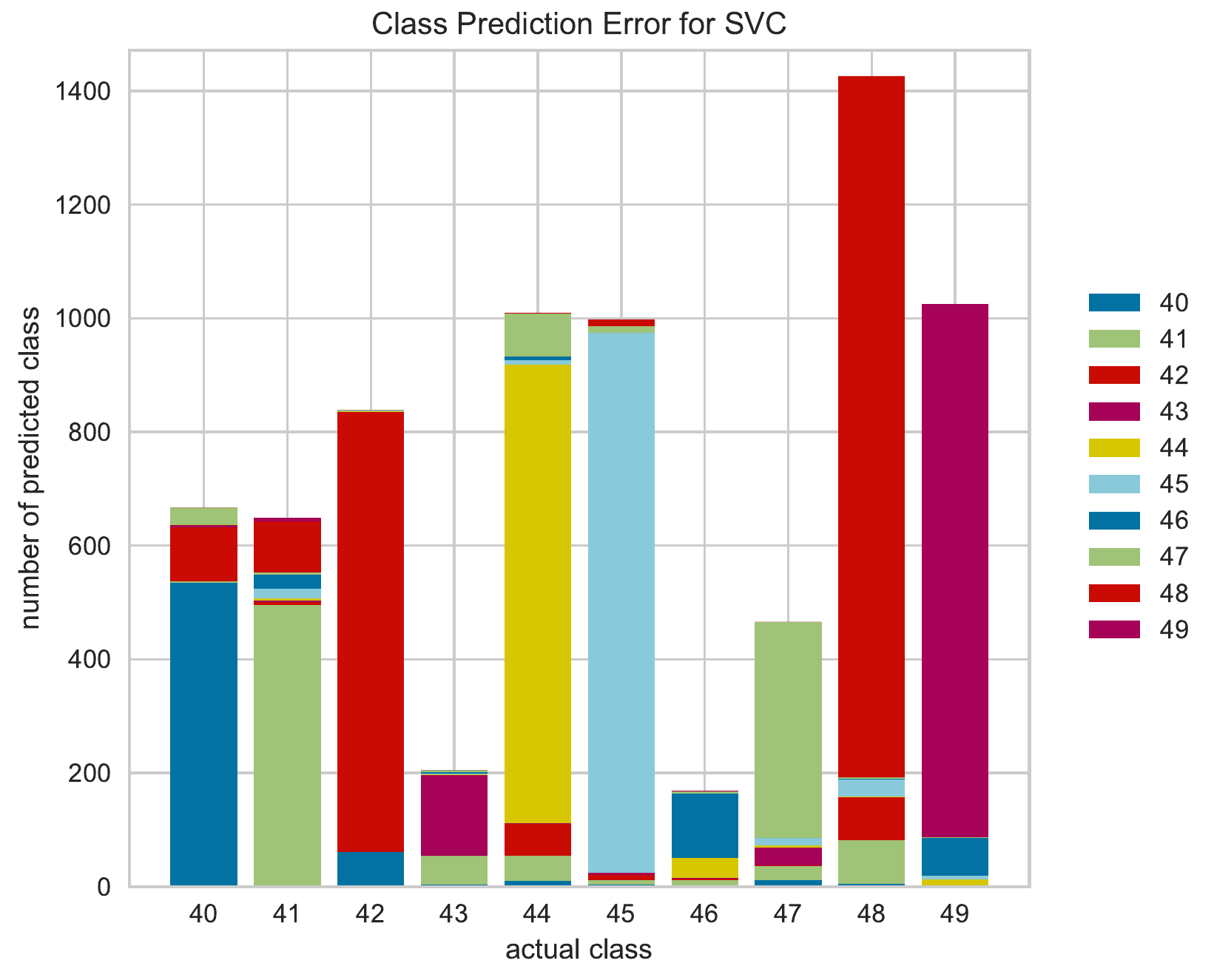}\\
  \caption{The classification results of 10 unseen classes on the PS of AWA2. The Confusion matrix (left) shows the results of predicted classes compared to their actual classes. In the class prediction error chart (right), the horizontal axis represents the actual class labels and the vertical axis represents the number of predicted classes.}\label{fig1and2}
\end{figure*}

\begin{figure*}
  \centering
  \includegraphics[scale=0.4]{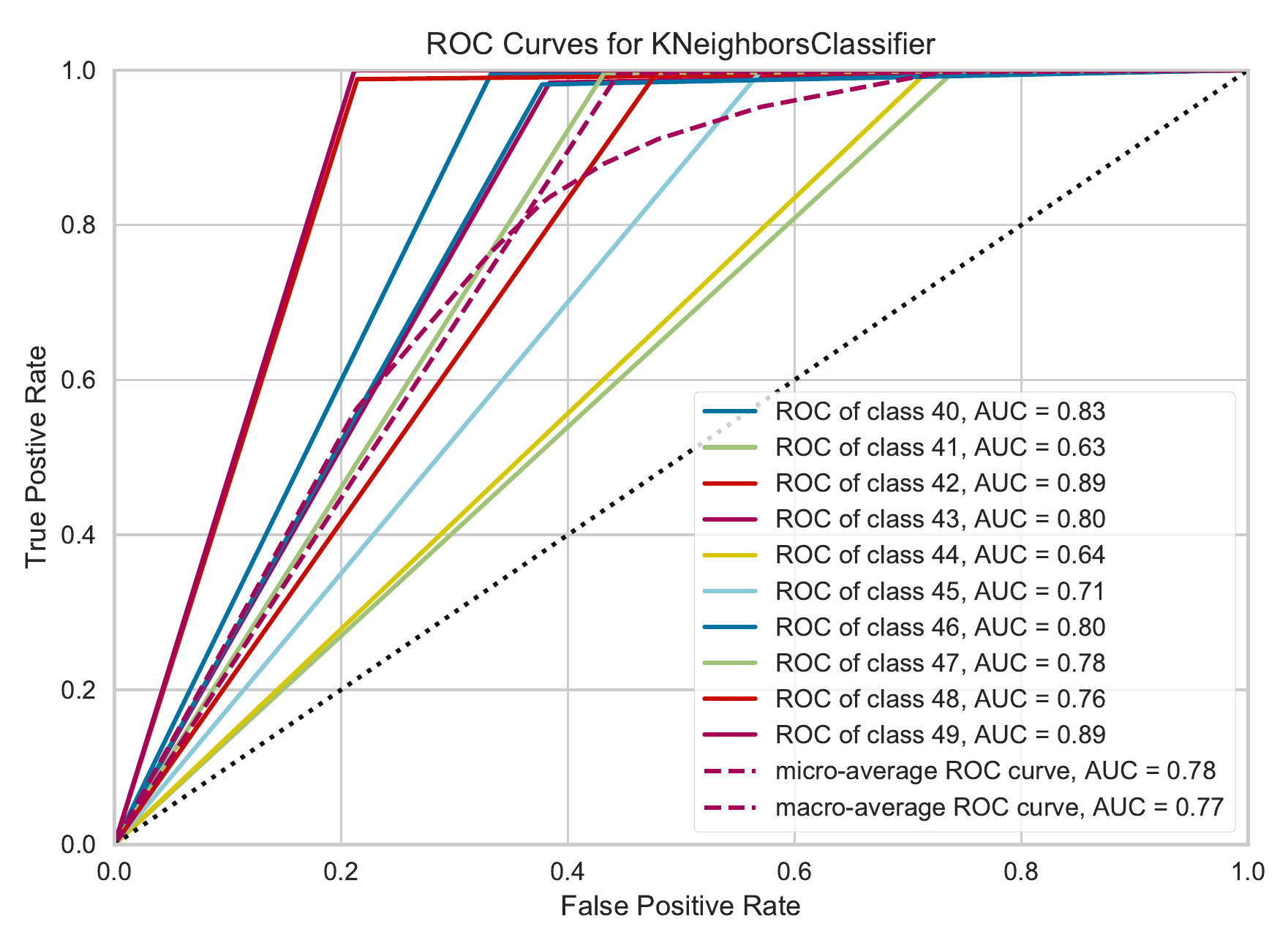}
  \includegraphics[scale=0.4]{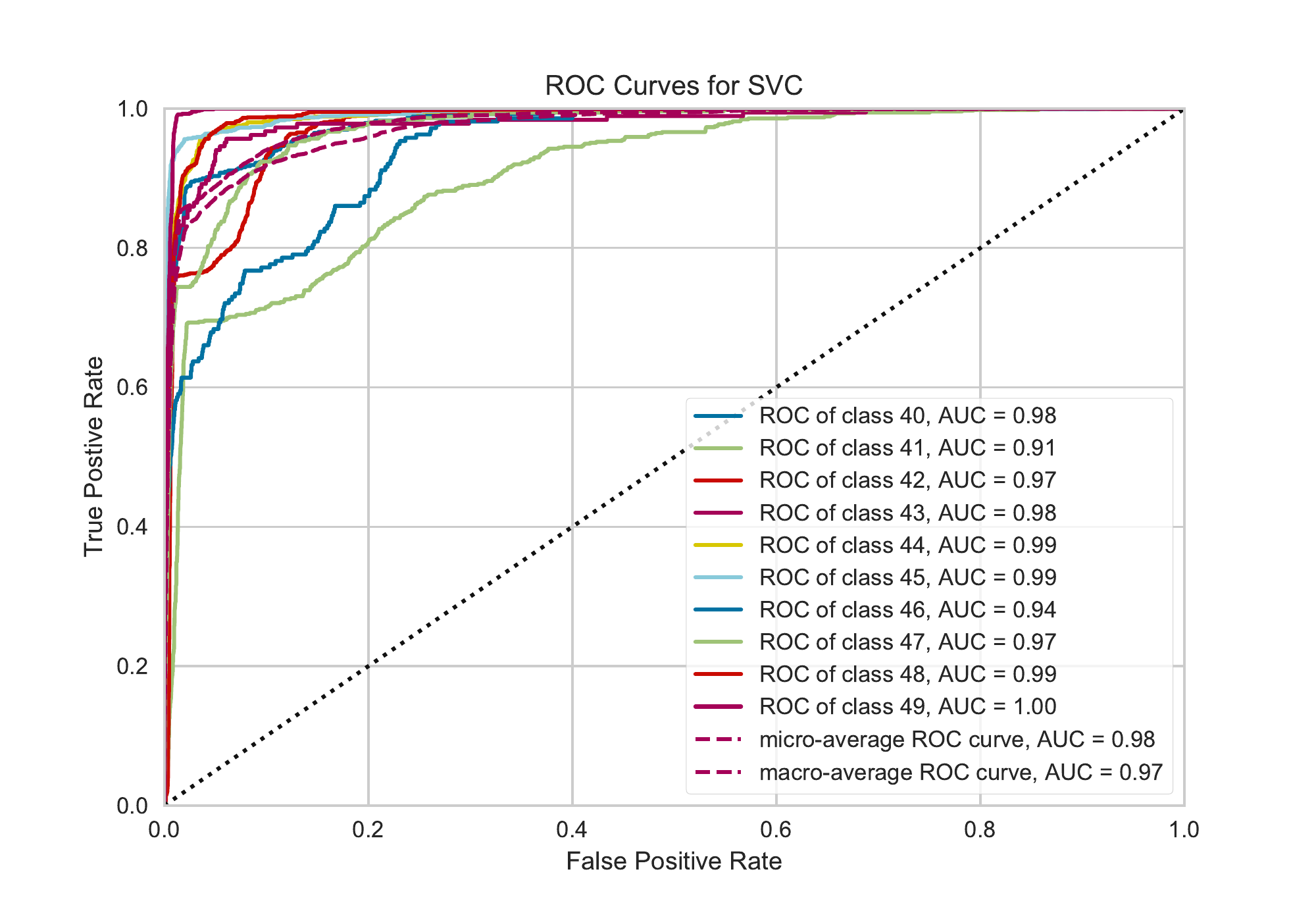}\\
  \caption{Comparing the ROC curve and the AUC value visualizations\cite{b38} for the KNN (left) and the SVM (right) classifiers that use the data generated from the same model setting.}\label{fig3and4}
\end{figure*}

Taking the PS of AWA2 as an example, Figure \ref{fig1and2} shows the classification results of 10 unseen classes. The average per class top-1 accuracy is 78.5\% shown in Table \ref{tab2}. The data on the diagonal of the confusion matrix indicates the correct number of classifications for each class. For example, there are 535 correct classifications of {\it sheep} and the false positives have 2 for {\it dolphins}, 95 for {\it bats}, 4 for {\it seals}, 1 for {\it blue+whale}, and 30 for {\it walrus}. Note that the class number 43 and 46 have relatively small number of correct in the confusion matrix with a high accuracy, as they have small total number in the dataset.

Different colors of the class prediction error chart from bottom to top represent the class number 40-49 in order, which can intuitively show the proportion of correct predictions of per class. Its vertical axis corresponds to the number of predicted classes in the left part, and it is observed that all the 10 unseen classes have a high accuracy.

The ROC curve and the AUC value visualize the tradeoff between the specificity (false positive rate) and the sensitivity (true positive rate) as a measure of the performance of the classifier\cite{b38}. Figure \ref{fig3and4} shows the results of the ROC curve and the AUC value for the KNN and the SVM classifier on the PS of AWA2. Each chart has 10 curves, and each curve represents the result classification of one class. Compared the two parts of Figure \ref{fig3and4}, we can find that the ROC curves of the 10 unseen classes in the right part are close to the top-left corner of the plot, where the AUC value is still higher than 0.9 for the lowest ROC of class 41. The ROC curve for the KNN classifier changes gently, i.e., the maximization of the true positive rate while minimizing the false positive rate\cite{b38} is smaller than the ROC curve of the SVM, and the AUC values are smaller than the AUC values of the SVM. The SVM classifier significantly performs better on our proposed model.

\subsection{The Generalized Zero-Shot Learning}
In real world applications, image classification problems do not have access to whether a novel image belongs to a seen or unseen class in advance. The generalized setting where both seen and unseen classes are used during testing is considered. Hence, generalized zero-shot learning is more meaningful from a practical point of view\cite{b4}.

Here, we use the same models trained on the PS of datasets and evaluate the performance of the generalized zero-shot learning, and details of the PS is showed in Table 1. The SVM is evaluated separately on both the seen classes and the unseen classes. We use the harmonic mean of the ${{\cal Y}_{\cal S}}$ and ${{\cal Y}_{\cal U}}$ accuracies\cite{b4} as a measure of evaluating the generalized zero-shot learning:
\begin{equation}
H = \frac{{2*ac{c_{{{\cal Y}_{\cal S}}}}*ac{c_{{{\cal Y}_{\cal U}}}}}}{{ac{c_{{{\cal Y}_{\cal S}}}} + ac{c_{{{\cal Y}_{\cal U}}}}}},
\end{equation}
where ${ac{c_{{{\cal Y}_{\cal S}}}}}$ and ${ac{c_{{{\cal Y}_{\cal U}}}}}$ represent the accuracy of the seen (${{\cal Y}_{\cal S}}$) and unseen (${{\cal Y}_{\cal U}}$) classes respectively.
\begin{table}
  \centering
  \caption{Generalized Zero-Shot Learning results on SUN, CUB, AWA1 and AWA2 using the PS split measuring the ${{\cal Y}_{\cal S}}$ and ${{\cal Y}_{\cal U}}$ top-1 accuracies. The results report the H(harmonic mean) in \%.}
  \begin{tabular}{c|c|c|c|c}
    \hline
  \textbf{Method}&	\textbf{SUN}&	\textbf{CUB}&	\textbf{AWA1}&	\textbf{AWA2}\\
  \hline
  DAP\cite{b12}&	7.2&	3.3&	0.0&	0.0\\
  IAP\cite{b12}&	1.8&	0.4&	4.1&	1.8	\\
  CONSE\cite{b8}&	11.6&	3.1&	0.8&	1.0\\
  CMT\cite{b15}&	11.8&	12.6&	1.8&	1.0	\\
  SSE\cite{b16}&	4.0&	14.4&	12.9&	14.8\\
  LATEM\cite{b6}&	19.5&	24.0&	13.3&	20.0\\
  ALE\cite{b13}&	26.3&	34.4&	27.5&	23.9\\
  DEVISE\cite{b35}&	20.9&	32.8&	22.4&	27.8\\
  SJE\cite{b36}&	19.8&	33.6&	19.6&	14.4\\
  ESZSL\cite{b14}&	15.8&	21.0&	12.1&	11.0\\
  SYNC\cite{b18}&	13.4&	19.8&	16.2&	18.0\\
  SAE\cite{b10}&	11.8&	13.6&	3.5&	2.2\\
      \hline
  \textbf{Ours}&	\textbf{48.3}&	\textbf{50.7}&	\textbf{76.1}&	\textbf{75.3}\\
    \hline
  \end{tabular}
\label{tab3}
\end{table}

The results for the generalized zero-shot learning is shown in Table \ref{tab3}. In the generalized zero-shot learning, our model accuracy is significantly improved on the SUN, CUB, AWA1 and AWA2, which are increased by 22\%, 19.3\%, 48.6\% and 47.5\%, respectively. On the coarse-grained datasets (AWA-1, AWA-2), the increase is larger than that on the fine-grained datasets (SUN, CUB).

As shown in Table \ref{tab3}, the generalized zero-shot learning results are lower than zero-shot learning results, this is due to the fact that training classes act as distractors for the image features that come from test classes\cite{b4}. Our proposed model can reconstruct the original image features and alleviate the problem of projection domain shift\cite{b9}. In the generalized zero-shot learning, more complicated techniques are necessary and our model studies the problem from a new perspective.

On the basis of the above, the discriminative embedding regulates the inter and intra class distances between the learned features and preserves the discriminative information. This clusters the same classes and separates the different classes, which benefits the learned features to be discriminative. The semantic embedding is used for generalizing the semantic knowledge from the seen classes to an unseen class. It joins in the intermediate embedding space, making the generator contains the image feature and the semantic information. The generator combines the semantic embeddings with the discriminative embeddings and utilizes the correlation of them to generate samples. The regressor feedback provides a generalization to the semantic space and the discriminative embedding space. The recurrent structure is trained to produce better samples at each iteration, realizing the coarse-to-fine process. This weakens the interference between seen and unseen classes and alleviates the susceptibility to domain shift. The significant improvement in the GZSL strongly suggests our proposed model is robust and universal.

\section{Conclusion and future work}
We propose a discriminative embedding autoencoder with a regressor feedback model for ZSL. The autoencoder is used for generating samples for classification of the unseen classes. For the classes-specific and high-quality unseen classes samples, we have two contributions about the models. The discriminative embedding regulates the inter/intra class distances between the learned features, which is learned from the image features. The encoder acts as the discriminator to learn the image features from high dimension to low dimension. The intermediate embedding space is jointly composed of the discriminative and semantic embedding space. The decoder aims to reconstruct the original image feature and provide a generalization to the unseen classes. The feedback mechanism allows the network to carry the reconstructed samples back to previous layers and refine the discriminative embedding and the semantic embedding. The recurrent structure is trained to produce better output at each iteration, realizing the coarse-to-fine process. The final goal of ZSL is to classify the unseen classes, and all the above operations are to generate better unseen classes samples, making the classifier more accurate. The experiment results show that our proposed model compares favorably with the state-of-the-art models on four benchmark datasets, especially the accuracy is significantly improved in the GZSL.

There are several improvements for the future work. The autoencoder can be replaced with any generative model such as GAN\cite{b39} or many variants as well. Exploration of more intricate forms of the attribute relations is used for classification of the unseen classes. The intermediate embedding space can fuse the multiple semantic representation of classes, etc.

\end{document}